\documentclass[journal]{IEEEtran}

\usepackage{times}
\usepackage{epsfig}
\usepackage{graphicx}
\usepackage{amsmath}
\usepackage{booktabs}
\usepackage{xcolor}
\usepackage{multirow}
\usepackage{amssymb}

\usepackage[hidelinks]{hyperref}
\hypersetup{
    colorlinks=true,%
    urlcolor=magenta
}

\makeatletter
\renewcommand{\maketag@@@}[1]{\hbox{\m@th\normalsize\normalfont#1}}%
\makeatother

\begin{document}

\title{Improving Generalization of Metric Learning via Listwise Self-distillation}

\author{Zelong~Zeng,
        Fan~Yang,
        Zheng~Wang,~\IEEEmembership{Member,~IEEE,}
        and~Shin'ichi~Satoh,~\IEEEmembership{Member,~IEEE}

\thanks{Z.~Zeng, F.~Yang, and S.~Satoh are with the Department of Information and Communication Engineering, Graduate School of Information Science and Technology, The University of Tokyo, Japan, and also with the Digital Content and Media Sciences Research Division, National Institute of Informatics, Japan. (e-mail: zzlbz@nii.ac.jp;yang@nii.ac.jp;satoh@nii.ac.jp)}
\thanks{Z.~Wang is with the School of Computer Science, National Engineering Research Center for Multimedia Software, Wuhan University, China. (e-mail: wangzwhu@whu.edu.cn)}
\thanks{This research is part of the results of Value Exchange Engineering, a joint research project between Mercari, Inc. and the RIISE.}
\thanks{Manuscript received XXX; revised XXX.}}

\markboth{ }%
{Zeng \MakeLowercase{\textit{et al.}}: Improving Generalization of Metric Learning via Listwise Self-distillation}

\maketitle

\begin{abstract}
Most deep metric learning (DML) methods employ a strategy that forces all positive samples to be close in the embedding space while keeping them away from negative ones. However, such a strategy ignores the internal relationships of positive (negative) samples and often leads to overfitting, especially in the presence of hard samples and mislabeled samples. In this work, we propose a simple yet effective regularization, namely Listwise Self-Distillation (LSD), which progressively distills a model’s own knowledge to adaptively assign a more appropriate distance target to each sample pair in a batch. LSD encourages smoother embeddings and information mining within positive (negative) samples as a way to mitigate overfitting and thus improve generalization. Our LSD can be directly integrated into general DML frameworks. Extensive experiments show that LSD consistently boosts the performance of various metric learning methods on multiple datasets.
\end{abstract}

\IEEEpeerreviewmaketitle
\section{Introduction}

Metric learning is widely used in vision tasks, where it is adopted to learn an embedding space that uses a specified distance metric to reasonably reflect the similarity between samples. A number of works have explored the effectiveness of metric learning in different vision tasks~\cite{schroff2015facenet,gordo2017end,brown2020smooth}.

Most deep metric learning methods employ a strategy that forces all positive samples to be close in the embedding space, while keeping them away from negative ones. To clarify, the definition of positive and negative samples is a relative concept. During training, a positive sample is defined as being in the same class as the anchor sample, and the opposite is a negative sample. Correspondingly, at the time of testing, samples of the same class as the query are positive, and vice versa. We call such a strategy \textit{hard embedding} as it only concerns the hard labels. However, \textit{hard embedding} tends to cause the following problems: (1) it makes the model ignore the internal information between samples, particularly, the similarity between intra-class samples. However, since the image retrieval task requires dealing with unseen classes, these internal information helps the model learn ``how to measure similarity'' and therefore ignoring it is detrimental to the generalization of the model. (2) The \textit{hard embedding} leads to overfitting, especially in the presence of hard samples and mislabeled samples. An interesting observation for deep models is that they can first adapt to easy samples, and gradually adapt to hard samples as the number of training epochs gets larger~\cite{han2018co}. Therefore, if hard samples are hard in nature, but we measure them with inappropriate metrics (\textit{e.g.}, the visual content of a sample is inherently less similar to other samples in the same class, or it was annotated with a wrong/noisy label, but we force its feature to be close with other intra-class samples), then the training will enter a phase of fitting hard samples with unreasonable features, which will directly lead to overfitting and poor generalization.

\begin{figure}[t]
\centering
\includegraphics[width=.85\linewidth]{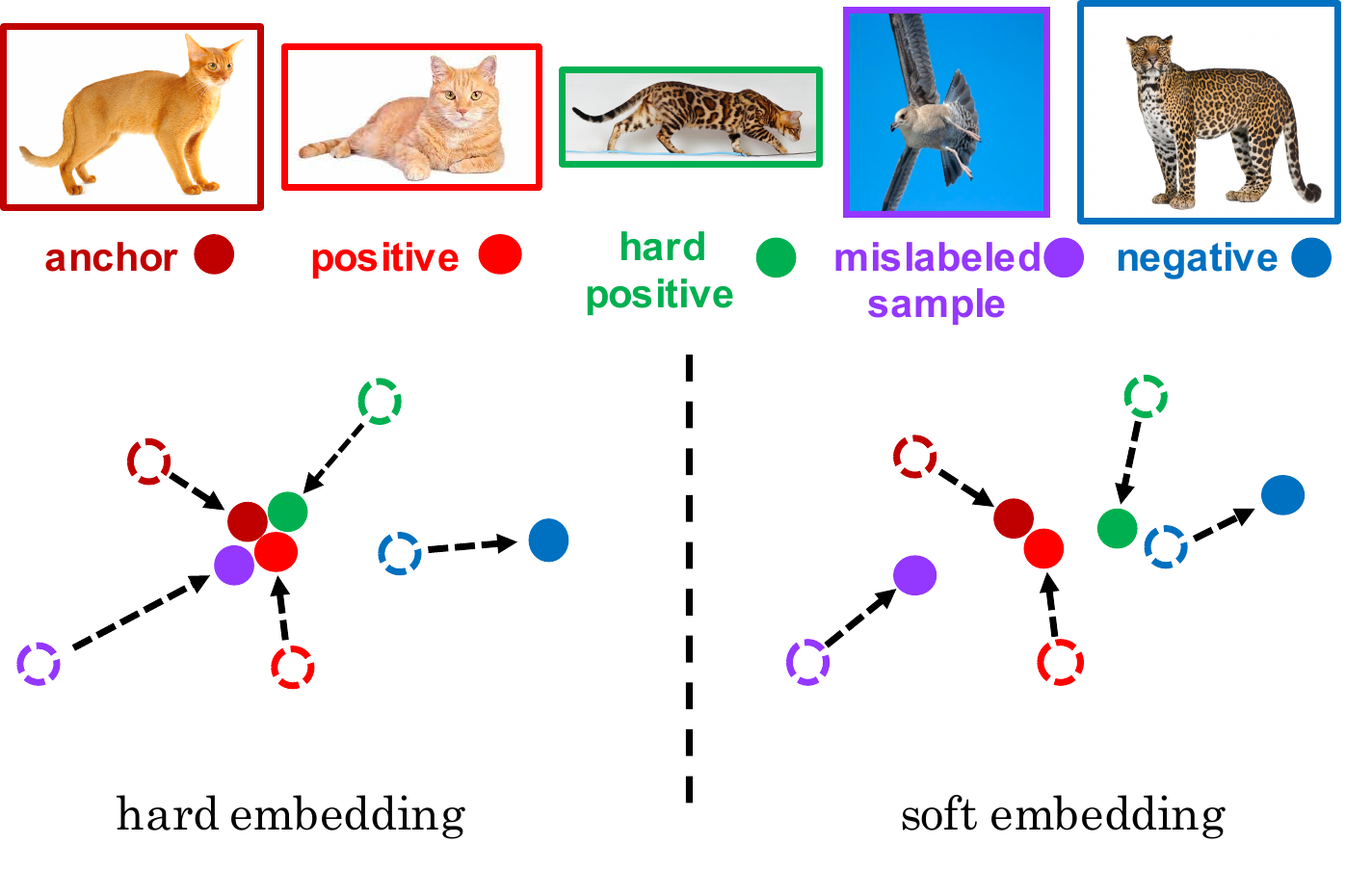}
\vspace{-2mm}
\caption{Toy example of \textit{hard embedding} vs \textit{soft embedding}: the \textit{hard embedding} (left) forces all hard positives and mislabeled samples to be close to the anchor, which causes the model to ignore the cat's texture and overfit. The \textit{soft embedding} (right) considers assigning a more appropriate distance target to each pair of samples. Compared with the easy positive, the hard positive (Bengal cat) is less similar to the anchor and the texture is more similar to the negative (leopard), so it should be slightly further away from the anchor and closer to the negative. The same is true for the mislabeled samples.}
\label{fig:embedding}
\end{figure}

In this paper, we propose a hypothesis that learning a smoother embedding space helps to improve the generalization of metric learning. That is, assigning a more appropriate distance target to each sample pair rather than just bringing samples in a positive pair closer and keeping samples in a negative pair apart based on hard labels. Fig.~\ref{fig:embedding} shows a toy example about \textit{hard embedding} and \textit{soft embedding}. To validate our hypothesis, inspired by the knowledge distillation~\cite{furlanello2018born,qin2021born}, we propose a simple yet effective regularization named listwise self-distillation (LSD) that distills the knowledge in a model itself and uses it to soften the \textit{hard embedding}, it means that the student model itself serves as the teacher model. The teacher uses its knowledge to measure each sample pair and provides more appropriate distance targets for the student. It is worth mentioning that LSD is a listwise method that has been shown computationally more efficient and more effective for ranking problems as it directly minimizes errors in ranking of samples~\cite{cao2007learning}. 

Our LSD can be directly integrated into a general DML framework. We demonstrate its effectiveness via a comprehensive set of evaluations with various metric learning methods on multiple datasets. We found that the intra-class compactness in the embedding space decreases as expected under the regularization of LSD, and the model performs significantly better than those without using LSD. In addition, we give theoretical explanations (Section~\ref{sec:theoretical}) on why LSD is effective, and show that LSD encourages the model to exploit more internal information between samples, especially for hard samples, allowing the model to better learn ``how to measure similarity''.

In summary, our contributions can be described as follows: 
\begin{itemize}
    \item We present the shortcomings of the existing metric learning methods on vision tasks that emphasizes only intra-class compactness and inter-class discrepancy.
    \item We propose a hypothesis that learning a smoother embedding space helps improve the generalization of metric learning, and validate our hypothesis on public image  retrieval benchmarks.
    \item We propose a simple yet effective regularization named Listwise Self-Distillation (LSD) for metric learning. LSD can be applied to any general DML framework. This is the first general approach to improve DML performance using knowledge self-distillation. We also provide theoretical explanations on why LSD works.
\end{itemize}

\section{Related Works}
\subsection{Metric Learning}

Metric learning are widely used to optimize networks by regulating the similarity between instances in the embedding space. Many metric learning methods use pair-based objectives defined on a tuple of samples, such as pairs~\cite{hadsell2006dimensionality}, triplets~\cite{wang2014learning,schroff2015facenet} or other variant like N-pairs~\cite{sohn2016improved} and lifted structure loss~\cite{oh2016deep}. Further, AP-based losses~\cite{brown2020smooth,rolinek2020optimizing} that directly optimising the Average Precision (AP) has been proven to be effective in metric learning. Moreover, some classification-based methods~\cite{deng2019arcface} and proxy-based methods~\cite{movshovitz2017no,qian2019softtriple} have shown the ability of learning distance-preserving embedding spaces.

Additionally, more involved research extending above objectives has been proposed: \cite{milbich2020diva} proposed and studies multiple complementary learning tasks; \cite{lin2018deep} proposed a framework to explicitly model the intra-class variance and disentangle the intra-class invariance; \cite{opitz2018deep} divides a large embedding into an ensemble of several smaller embeddings and \cite{kim2018attention} proposed an attention-based ensembling framework.

Due to the large variety of methods for metric learning, it is inefficient to design a regularization method for each method. Our motivation is to propose a generalized regularization method for metric learning.
Considering that metric learning methods all require a backbone model to extract features from the input data, usually images, LSD is designed based on such features and is therefore applicable to various metric learning methods (as shown in Section~\ref{sec:ablation}). 

\noindent \textbf{Remarks.} Here, we would like to highlight the difference of motivation between our LSD and some similar works \cite{milbich2020diva, lin2018deep}. \cite{lin2018deep} also proposed that we need to focus on intra-class variance, but it ignore the inter-class variance. Our ``soft embeddings'' imply not only ``intra-class'' structures but also ``inter-class'' structures. \cite{milbich2020diva} applies multiple complementary learning tasks for learning multiple relation between different samples and result in strong generalization, but all learning tasks are set manually, while our LSD applies self-distillation to set the learning target by the model itself.

\subsection{Knowledge Distillation}

Conventional Knowledge Distillation (KD)~\cite{hinton2015distilling,zagoruyko2016paying,tung2019similarity} methods aim at model compression: they use knowledge from the teacher model to generate compact models, achieving a good performance-efficiency trade-off. Further, some works has found that the student model can outperform the teacher model if the student is configured with the same capacity as its teacher. Born Again Network~\cite{furlanello2018born} proposed to use pre-trained models as teacher models to supervise the training of student models. \cite{zhang2018deep} proposed deep mutual learning in which an ensemble of students learn collaboratively and teach each other throughout the training process. PS-KD~\cite{kim2021self}, BYOT~\cite{zhang2019your} and CS-KD~\cite{yun2020regularizing} proposed the self-distillation where students becoming teachers themselves. \cite{ge2021self} proposed to apply random-walk to refine the output of model as refined learning target. These works produced student models that achieved better accuracy on classification problems. More recently, some works start to focus attention on retrieval task: A novel work~\cite{qin2021born}, named Born Again neural Rankers (BAR), applied born-again-based knowledge distillation technology on learning to rank (LTR) task. Another work~\cite{roth2021simultaneous}, called Simultaneous Similarity-based Self-distillation (S2SD), extends DML with knowledge distillation from auxiliary, high-dimensional embedding and feature spaces to leverage complementary context during training while retaining test-time cost.

\noindent \textbf{Remarks.} Here, we would like to highlight the marked difference between our LSD and most existing KD methods. First, most existing knowledge distillation methods, either for model compression or performance improvement, focus on classification tasks, while our LSD is designed for metric learning. Second, these approaches typically use teachers to predict the probability distribution over all classes and then use this ``dark knowledge'' to enhance students. Therefore, these methods require classifiers or memory to output the probability distribution and are not applicable to most metric learning methods. Since our LSD does not require an additional classifier or memory module, it is more general.

Besides, the closely related works are BAR~\cite{qin2021born} and S2SD~\cite{ge2021self}: 1). BAR focuses on the learning to rank (LTR) problem and only evaluated their method on some recommender system datasets (\textit{i.e.}, web search datasets). However, our LSD focuses on metric learning and can work on end-to-end CNN frameworks. Our LSD also employs a self-distillation structure, which can greatly improve the training efficiency compared with the born-again structure. 2). Compared with S2SD, our motivation is different. 
S2SD targets the task of ``compression'', \textit{i.e.}, using high-dimensional embedding space with better generalization capacity to assist the training of low-dimensional embedding space. In contrast, The motivation of LSD is to reduce the potential overfitting risk of hard labels by self-distillation, i.e., by creating ``soft embeddings''. Our LSD proposes that the same dimensional embedding (as opposed to the high-dimensional embedding of S2SD) space itself can provide useful information to improve performance, while S2SD DOES NOT. 

\begin{figure*}[t]
\centering
\includegraphics[width=0.9\linewidth]{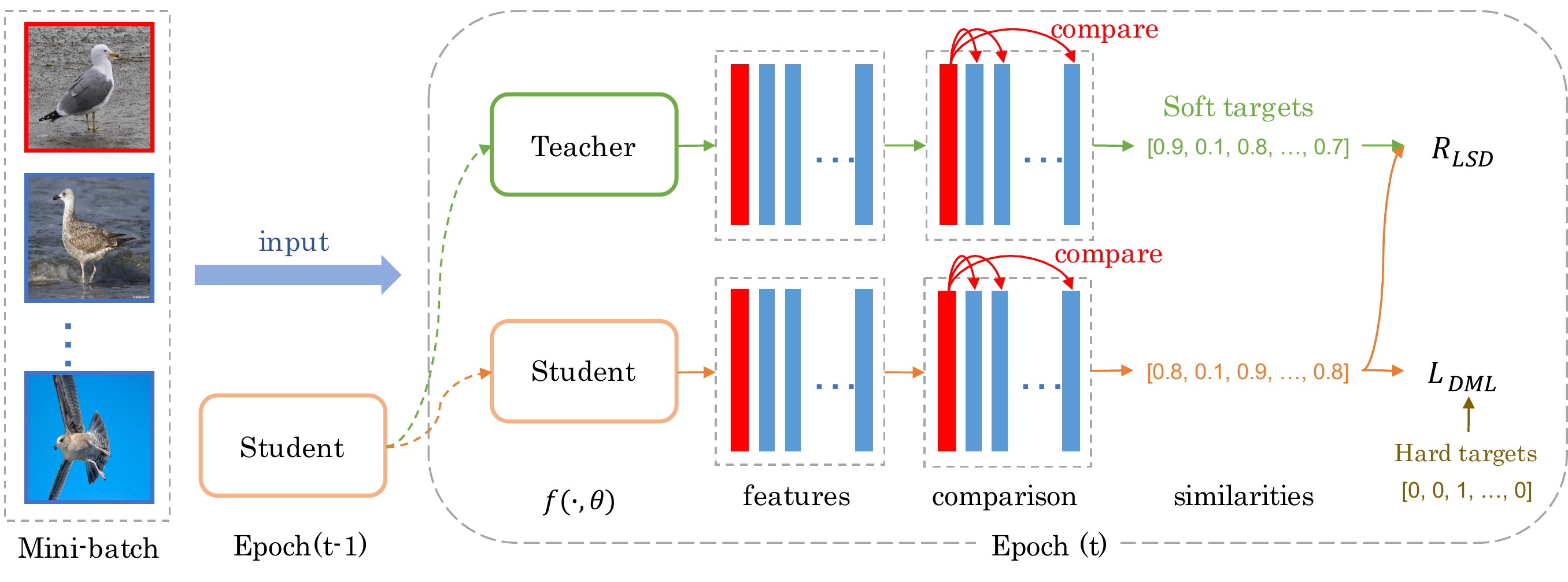}
\caption{A schematic diagram of LSD. The student at epoch $t-1$ is the teacher at epoch $t$. The student is trained with LSD regularization and any DML loss. The teacher predicts the similarity scores between an anchor (red) and its positives/negatives (blue) in a mini-batch as the soft targets for computing the LSD regularization. DML loss is computed by using hard labels.}
\label{fig:framework}
\end{figure*}

\section{Methodology}

\subsection{Problem Formulation}
For metric learning, the training data can be donated as a set $\mathcal{X} = \left \{ \boldsymbol{x}_{1},\boldsymbol{x}_{2},...,\boldsymbol{x}_{n} \right \}$, $y_i$ is the corresponding label of $\boldsymbol{x}_i$. Our goal is to learn an embedding function $f:\mathcal{X} \mapsto \Phi \subset \mathbb{R}^d$ that maps each input $\boldsymbol{x}\in\mathcal{X}$ to the $d$-dimensional embedding space $\Phi$, which allows to measure the similarity between inputs. To this end, let $f:=f(\cdot,\theta)$ be a differentiable deep network with parameters $\theta$, whose output $\boldsymbol{v}_{i} = f(\boldsymbol{x}_{i},\theta)$ is typically normalized to $\boldsymbol{z}_{i} = \frac{\boldsymbol{v}_{i}}{\left \| \boldsymbol{v}_{i} \right \|}$. The distance (or similarity) between two inputs $\boldsymbol{x}_{i}, \boldsymbol{x}_{j}$ is defined as $D_{ij}:=d(\boldsymbol{z}_{i},\boldsymbol{z}_{j})$, where $d(\cdot,\cdot)$ denotes a pre-defined distance (or similarity) function, \textit{e.g.}, Euclidean distance.

\subsection{Listwise Knowledge Distillation}
Since most of the previous KD methods focus on classification tasks, we need to design a KD function that is applicable to metric learning. In general, the KD function is based on the original objective function. However, since different metric learning methods have different learning objectives, accommodating various learning objectives will lead us to tailor the KD function according to different metric learning methods, which is not conducive to the generality of the KD function and contradicts our goal of proposing a generic KD function.

To propose a generic KD function, it needs to be designed based on functions that are applicable to most metric learning methods. Inspired by previous works~\cite{cao2007learning,zagoruyko2016paying,qin2021born}, we chose the Listwise Softmax Cross Entropy loss (LSCE) which is widely used in LTR (Learning to Rank) tasks: 
\begin{equation}
    L_{\textup{Softmax}}(\mathbf{y},\mathbf{s}) = -\sum_{i=1}^{n}y_i \textup{ln}\frac{\textup{exp}(s_i)}{\sum_{j=1}^{n}\textup{exp}(s_j)},
    \label{eq:original_softmax}
\end{equation}
where $\mathbf{y}$ and $\mathbf{s}$ are the list of ranking labels and scores, respectively. Especially, $\mathbf{y}_i$ indicates the probability that the $i$-th sample is ranked first in the batch in terms of similarity to the query. LSCE requires only features and labels, which makes it applicable to most deep learning networks. In addition, as a listwise approach, it has advantages in terms of computational efficiency and effectiveness in ranking problems.
Inspired by LSCE, we propose listwise knowledge distillation function. Within a mini-batch $\mathcal{B}\subset \mathcal{X}$ with an indexing set $\mathcal{B}^\textrm{ID}$, let $i\in\mathcal{B}^\textrm{ID}$ be the index of an arbitrary sample, and $j(k)$ be the index of other sample in the same mini-batch, our listwise knowledge distillation function can be described as: 
\begin{equation}
    R_{\textup{KD}}(\mathcal{B}) = \frac{-1}{\left | \mathcal{B} \right |^2}
    \displaystyle\sum_{i \in \mathcal{B}^{\textrm{ID}}}
    \displaystyle\sum_{j\in\mathcal{B}^{\textrm{ID}}}\sigma(s_{ij}^{(T)}/\tau)\textup{ln}(\sigma(s_{ij}^{(S)}/\tau)),
    \label{eq:listwise_KD}
\end{equation}
where $\sigma()$ indicates the softmax function:
\begin{equation}
    \sigma(s_{ij}/\tau)=\textup{ln}\frac{\textup{exp}(s_{ij}/\tau)}{\sum_{k\in\mathcal{B}^{\textrm{ID}}}\textup{exp}(s_{ik}/\tau)},
    \label{eq:softmax}
\end{equation}
Here, $\tau$ is a temperature parameter, $s^{(T)}_{ij}$ and $s^{(S)}_{ij}$ are the similarity between $\boldsymbol{x}_{i}$ and $\boldsymbol{x}_{j}$ predicted from teacher model $f(\boldsymbol{x}_{i},\theta_{T})$ and student model $f(\boldsymbol{x}_{i},\theta_{S})$, respectively. We use softmax function to convert the similarity $s^{(T)}_{ij}$ into ranking probability, and depending on the predefined distance function in the metric learning method, the $s_{ij}$ can be formulated as:
\begin{equation}
    s_{ij}=\begin{cases}
        d(\boldsymbol{z}_{i},\boldsymbol{z}_{j}) & \text{ if } d(\boldsymbol{z}_{i},\boldsymbol{z}_{j})=\boldsymbol{z}_{i} \cdot \boldsymbol{z}_{j} \\ 
        1-\frac{d(\boldsymbol{z}_{i},\boldsymbol{z}_{j})^2}{2} & \text{ if } d(\boldsymbol{z}_{i},\boldsymbol{z}_{j})= \left \| \boldsymbol{z}_{i}-\boldsymbol{z}_{j} \right \|_2,
    \end{cases}
\label{eq:similarity}
\end{equation}
where $\boldsymbol{z}_{i}$ is the unit feature vector of input $\boldsymbol{x}_{i}$ projected from the model $f(\boldsymbol{x}_{i},\theta)$, the $\cdot$ operation denotes the inner (dot) product and $\left \| \cdot \right \|_2$ donates the Euclidean norm. 

\subsection{Listwise Self-distillation}
\label{sec:LSD}
We propose a new KD approach, called listwise self-distillation (LSD), which distills the model's own knowledge to improve its generalization capability. Specifically, a student model is its own teacher and utilizes its past measurements to provide more informative supervisions during training. Fig.~\ref{fig:framework} shows the framework of LSD. Let $s^{(t)}_{ij}$ be the similarity between $i$-th and $j$-th samples in a batch measured by the model at $t$-th epoch, our LSD objective at $t$-th epoch can be written as:
\begin{equation}
    R_{\textup{LSD}}(\mathcal{B}) = \frac{-1}{\left | \mathcal{B} \right |^2}
    \displaystyle\sum_{i \in \mathcal{B}^{\textrm{ID}}}
    \displaystyle\sum_{j\in\mathcal{B}^{\textrm{ID}}}\alpha\sigma(s_{ij}^{(T)}/\tau)\textup{ln}(\sigma(s_{ij}^{(S)}/\tau)),
    \label{eq:lsd}
\end{equation}
where $\alpha$ is the weight for soft targets. In the conventional KD, the teacher is a reliable pre-trained model, so the $\alpha$ is usually set to a fixed value during training. However, for our self-distillation model, the model does not learn enough knowledge in the early epochs of training, but as the number of epochs grows, the model becomes increasingly reliable. Thus, as in PS-KD~\cite{kim2021self}, we set the value of $\alpha$ to be dynamic:
\begin{equation}
    \alpha_t = \frac{t}{T},
    \label{eq:alpha_t}
\end{equation}
where $\alpha_t$ is the value of $\alpha$ at the $t$-th epoch and $T$ is the total epoch for training. The value of $\alpha$ gradually increases along with the number of epochs, \textit{i.e.}, as the reliability of the model itself increases. We also try to use other functions to set the value of $\alpha$, \textit{e.g.}, exponentially. However, we found that a simple linear growth can achieve the best results. To summarize, our LSD function at $t$-th epoch can be described as:
\begin{equation}
    R_{\textup{LSD}}(\mathcal{B}) = \frac{-1}{\left | \mathcal{B} \right |^2}
    \displaystyle\sum_{i \in \mathcal{B}^{\textrm{ID}}}
    \displaystyle\sum_{j\in\mathcal{B}^{\textrm{ID}}}\frac{t}{T}\sigma(s_{ij}^{(T)}/\tau)\textup{ln}(\sigma(s_{ij}^{(S)}/\tau)),
\label{eq:lsd2}
\end{equation}

Overall, the objective function for training is:
\begin{equation}
    \mathcal{L} = L_{\textrm{DML}} + \tau^2\lambda R_{\textrm{LSD}},
\label{eq:lsd3}
\end{equation}
where $L_{\textrm{DML}}$ is an arbitrary metric learning loss function and $\lambda$ is the weight of LSD.

\section{Theoretical Support}
\label{sec:theoretical}
In this section, we analyze the gradients of listwise self-distillation loss to give theoretical support for the reason why LSD works. First, the gradient for $R_{\textrm{LSD}}$ (for convenience, $R_{\textrm{LSD}}$ here is for an anchor $\boldsymbol{x}_{i}$) with respect to the embedding $\boldsymbol{v}_i$ at $t$-th epoch has the following form\footnote{We provide a full derivation of the loss gradient in the Appendix.~\ref{appendix_gradient}}: 
\begin{equation}
        \begin{split}
        & \left. \frac{\partial R_{LSD}}{\partial \boldsymbol{v}_i} \right|_{\mathcal{P}(i)} \\
        & = \frac{t}{\tau T \left | \mathcal{B} \right | \left \| \boldsymbol{v}_i \right \|}\sum_{p \in \mathcal{P}(i)} \left ( \boldsymbol{z}_p -\left ( \boldsymbol{z}_i \cdot \boldsymbol{z}_p \right ) \boldsymbol{z}_i\right ) \left (  P_{ip} - S_{ip}\right )
        \end{split}
    \label{eq:positive_grad}
\end{equation}

\begin{equation}
        \begin{split}
        & \left. \frac{\partial R_{LSD}}{\partial \boldsymbol{v}_i} \right|_{\mathcal{N}(i)} \\
        & = \frac{t}{\tau T \left | \mathcal{B} \right | \left \| \boldsymbol{v}_i \right \|}\sum_{n \in \mathcal{N}(i)} \left ( \boldsymbol{z}_n -\left ( \boldsymbol{z}_i \cdot \boldsymbol{z}_n \right ) \boldsymbol{z}_i\right ) \left (  P_{in} - S_{in}\right )
        \end{split}
    \label{eq:negative_grad}
\end{equation}
Here, $\mathcal{P}(i)$ and $\mathcal{N}(i)$ are the positive set and negative set of $\boldsymbol{x}_{i}$ in the mini-batch $\mathcal{B}$, respectively. $\boldsymbol{z}_{i}$ and $\boldsymbol{z}_{i}^{(T)}$ are unit feature vector of input $\boldsymbol{x}_i$ from student model and teacher model, respectively.
$P_{ik} = \frac{ \textup{exp} \left( \boldsymbol{z}_i \cdot \boldsymbol{z}_k / \tau \right)}{\sum_{k \in \mathcal{B}^{\textrm{ID}}}\textup{exp} \left( \boldsymbol{z}_i \cdot \boldsymbol{z}_k / \tau \right)} $ and 
$S_{ik} = \frac{ \textup{exp} \left( \boldsymbol{z}_i^{(T)} \cdot \boldsymbol{z}_k^{(T)} / \tau \right)}{\sum_{k \in \mathcal{B}^{\textrm{ID}}}\textup{exp} \left( \boldsymbol{z}_i^{(T)} \cdot \boldsymbol{z}_k^{(T)} / \tau \right)} $. 
Then, we show that LSD focuses more on hard samples. Here, taking easy and hard positive samples as examples, for an easy positive sample $\boldsymbol{z}_{i} \cdot \boldsymbol{z}_{p} \approx 1$:
\begin{equation}
    \footnotesize
    \begin{split}
        \|\boldsymbol{z}_{p}-\left(\boldsymbol{z}_{i} \cdot \boldsymbol{z}_{p}\right) \boldsymbol{z}_{i} \|=\sqrt{1-\left(\boldsymbol{z}_{i} \cdot \boldsymbol{z}_{p}\right)^{2}} 
        \approx 0,
    \end{split}
    \label{eq:easy_samples}
\end{equation}
However, for an hard positive sample $0 < \boldsymbol{z}_{i} \cdot \boldsymbol{z}_{p} \ll 1$:
\begin{equation}
    \footnotesize
    \begin{split}
        \|\boldsymbol{z}_{p}-\left(\boldsymbol{z}_{i} \cdot \boldsymbol{z}_{p}\right) \boldsymbol{z}_{i} \|=\sqrt{1-\left(\boldsymbol{z}_{i} \cdot \boldsymbol{z}_{p}\right)^{2}} 
        \approx 1,
    \end{split}
    \label{eq:hard_samples}
\end{equation}
Thus, for easy samples, the gradient of $R_{\textrm{LSD}}$ along $\boldsymbol{v}_i$ is close to $0$, while for hard samples:
\begin{equation}
    \footnotesize
    \begin{split}
        &\|\boldsymbol{z}_{p}-\left(\boldsymbol{z}_{i} \cdot \boldsymbol{z}_{p}\right) \boldsymbol{z}_{i} \|\left|P_{i p}-S_{i p}\right|\\
        &\approx \left|P_{i p}-S_{i p}\right|,
    \end{split}
    \label{eq:grad_norm}
\end{equation}

We found that gradient contribution of easy samples is smaller (Eq~\ref{eq:easy_samples}) while the gradient contribution of hard samples is larger (Eq~\ref{eq:hard_samples}), which implies that LSD gives more weight to hard samples during training. Moreover, $P_{ip}$/$P_{in}$ is the listwise similarity from the student and $S_{ip}$/$S_{in}$ is that from the teacher. For those samples with large differences in output between the teacher and the student, \textit{i.e.}, those samples that do not fit well in the network, $R_{\textrm{LSD}}$ produces larger gradients (Eq~\ref{eq:grad_norm}). This means that LSD encourages the model to focus on those samples that are poorly fitted and to mine deeper information form samples, especially for those that are difficult to fit. At the same time, this part will also use distilled knowledge to assign more appropriate targets to hard samples. Thus reducing the risk of overfitting caused by hard labels or noisy labels (Section~\ref{subsubsec:soft embedding}).

\tabcolsep=5pt
\begin{table*}
\caption{Comparison of Recall@1, Recall@10 and mAP for all metric learning methods. Each model is trained using the same training setting over 150 epochs for CUB/CARS and 100 epochs for SOP. The notation ‘+LSD’ means that the indicated model is trained with LSD. The `(R)', `(H)' and `(D)' suffix denote the random, semi-hard and distance-based sampling strategy, respectively. \textbf{Bold} denotes the best result among using LSD and not using LSD with the same metric learning loss function function. \textbf{\textcolor{blue}{Boldblue}} marks overall best results.}
\centering
\resizebox{\linewidth}{!}{
    \begin{tabular}{l|cc|cc|cc}
    \toprule
    Benchmarks & \multicolumn{2}{c}{CUB200-2011} & \multicolumn{2}{c}{CARS196} & \multicolumn{2}{c}{SOP} \\
    \midrule
    Approaches$\rightarrow$ & \textbf{R@1}  & \textbf{mAP}  & \textbf{R@1}  & \textbf{mAP}  & \textbf{R@1}  & \textbf{mAP} \\
    \midrule
    Contrastive (D)         & 60.18         &32.65 & 75.64& 31.64& 72.86& 41.13\\
    Contrastive (D) + LSD (Ours)   
    & \textbf{61.63 ($\uparrow$ 1.45)}& \textbf{32.88 ($\uparrow$ 0.23)}& \textbf{76.76 ($\uparrow$ 1.12)}& \textbf{32.42 ($\uparrow$ 0.78)}& \textbf{73.58 ($\uparrow$ 0.72)}& \textbf{42.05 ($\uparrow$ 0.92)}\\
    \midrule
    SmoothAP            & 59.98& 31.23& 75.57& 30.42& 75.76& 43.41\\
    SmoothAP + LSD (Ours)      & \textbf{61.53($\uparrow$ 1.55)}&  \textbf{32.62($\uparrow$ 1.39)}& \textbf{77.74($\uparrow$ 2.17)}&  \textbf{32.96($\uparrow$ 2.54)}& \textbf{76.79($\uparrow$ 1.03)}&  \textbf{44.53($\uparrow$ 1.12)}\\
    \midrule
    ProxyNCA            & \textbf{63.12}&  31.73& 79.04&  31.51& -&  -\\
    ProxyNCA + LSD (Ours)      & 62.22($\downarrow$ 0.90)&  \textbf{32.86($\uparrow$ 1.13)}& \textbf{79.76($\uparrow$ 0.72)}&  \textbf{32.83($\uparrow$ 1.32)}& -&  -\\
    \midrule
    Triplet (R)             & 58.25&  29.83& 69.18& 27.65& 66.75& 33.77\\
    Triplet (R) + LSD (Ours)       & \textbf{59.59($\uparrow$ 1.34)}& \textbf{31.36($\uparrow$ 1.53)}& \textbf{73.10($\uparrow$ 3.92)}&  \textbf{29.81($\uparrow$ 2.16)}& \textbf{67.60($\uparrow$ 0.85)}& \textbf{34.90($\uparrow$ 1.13)}\\
    Triplet (H)             & 59.67& 31.13& 71.93& 29.58& 73.57& 41.17 \\
    Triplet (H) + Pointwise BAR           & 59.27($\downarrow$ 0.40)&  31.09($\downarrow$ 0.04)& 71.91($\downarrow$ 0.02)&  29.38($\downarrow$ 0.20)& 73.60($\uparrow$ 0.03)& 41.14($\downarrow$ 0.03)\\
    Triplet (H) + LSD (Ours)       & \textbf{60.92($\uparrow$ 1.25)}&  \textbf{32.67($\uparrow$ 1.54)}& \textbf{75.05($\uparrow$ 3.12)}&  \textbf{31.79($\uparrow$ 2.21)}& \textbf{74.70($\uparrow$ 1.13)}&  \textbf{42.11($\uparrow$ 0.94)} \\
    Triplet (S)             & 61.34& 32.09& 77.36& 31.22& 73.23& 40.57\\
    Triplet (S) + LSD (Ours)       & \textbf{61.74($\uparrow$ 0.40)}&  \textbf{32.37($\uparrow$ 0.28)}& \textbf{78.40($\uparrow$ 1.04)}&  \textbf{33.28($\uparrow$ 1.06)}& \textbf{73.75($\uparrow$ 0.52)}&  \textbf{41.22($\uparrow$ 0.65)} \\
    Triplet (D)             & 63.15& 32.56& 78.10& 32.53& 77.13& 45.17\\
    Triplet (D) + LSD (Ours)       & \textbf{64.38($\uparrow$ 1.23)}&  \textbf{33.17($\uparrow$ 0.61)}& \textbf{79.25($\uparrow$ 1.15)}&  \textbf{33.62($\uparrow$ 1.09)}& \textbf{78.18($\uparrow$ 1.05)}&  \textbf{46.16($\uparrow$ 0.99)} \\
    \midrule
    Margin (D)              & 62.85&  32.24& 80.01& 33.02& 78.23& 46.51\\
    Margin (D) + LSD (Ours) & \textbf{\textcolor{blue}{63.79($\uparrow$ 0.94)}}&  \textbf{\textcolor{blue}{33.21($\uparrow$ 0.97)}}& \textbf{80.68($\uparrow$ 0.67)}& \textbf{\textcolor{blue}{34.14($\uparrow$ 1.12)}}& \textbf{\textcolor{blue}{78.87($\uparrow$ 0.64)}}& \textbf{\textcolor{blue}{47.17($\uparrow$ 0.66)}}\\
    \midrule
    Multisimilarity              & 62.39&  31.37& 81.27& 31.99& 77.54& 45.31\\
    Multisimilarity + LSD (Ours) &\textbf{63.76($\uparrow$ 1.37)}&  \textbf{32.11($\uparrow$ 0.74)}& \textbf{\textcolor{blue}{82.27($\uparrow$ 1.00)}}& \textbf{32.65($\uparrow$ 0.66)}& \textbf{78.50($\uparrow$ 0.96)}& \textbf{46.40($\uparrow$ 1.09)}\\
    \midrule
    PS-KD                
    & 58.86& 30.38& 73.79& 29.43& 65.67& 20.28\\
    CS-KD                   
    & 59.06&  30.38& 73.62&  29.63& 48.91&  19.87\\

    \bottomrule
    \end{tabular}
    }
\label{tab:results}
\end{table*}

\section{Experiments}
\subsection{Datasets} 
We use CUB200-2011~\cite{wah2011caltech}, CARS196~\cite{krause20133d} and Stanford Online Products (SOP)~\cite{oh2016deep} as benchmarking datasets:

\noindent \textbf{CUB200-2011} contains $11,788$ images in $200$ classes of birds. Follwing~\cite{wah2011caltech}, Train/Test sets are made up of the first/last $100$ classes ($5,864/5,924$ images respectively). Samples are distributed evenly across classes.

\noindent\textbf{CARS196} contains $16,185$ images of $196$ car categories with even sample distribution. Follwing~\cite{krause20133d}, Train/Test sets use the first/last $98$ classes ($8054/8131$ images).

\noindent \textbf{SOP} contains $120,053$ online product images in $22,634$ categories. Train/Test sets are provided, contain $59,551$ images of $11,318$ classes in the Train set and $60,502$ images of $11,316$ classes in the Test set~\cite{oh2016deep}. In SOP, there are only 2 to 10 images for each category, leading to significantly different data distribution compared to CUB200-2011 and CARS196.

\subsection{Experimental Protocol}
We follow the training protocol in~\cite{roth2020revisiting} for fair comparison. Specifically, we utilize a ResNet50 with frozen Batch-Normalization layers as the backbone network. The weights of the backbone were pre-trained on the ImageNet dataset. A 128-dimensional fully-connected layer is added after the global pooling layer, as the default embedding dimension is set to 128. Following standard practice, we randomly resize and crop images to $224 \times 224$ for training, and center crop them to the same size at evaluation.  Random horizontal flipping ($p=0.5$) is also utilized in training.  In all experiments, we use the Adam optimizer with a fixed learning rate of $10^{-5}$ and a weight decay of $4 \times 10^{-4}$. Note that there is no learning rate scheduling for unbiased comparison. In addition, LSD requires two hyper-parameters. We set the distillation regularization weight $\lambda$ to 500 for CUB200-2011, 75 for CARS196 and 100 for Stanford Online Products, and set the temperature to 1 by default (More detailed experiments and discussions on hyper-parameters, please refer to Section~\ref{sec:ablation})
All experiments were performed on a single Nvidia V100 GPU. Each training was run over $150$ epochs for CUB200-2011/CARS196 and $100$ epochs for Stanford Online Products, if not mentioned. In evaluation, we use Recall@$1$ and mean Average Precision (mAP) as evaluation metrics.

\subsection{Ablation Study}
\label{sec:ablation}
We provide ablation study on multiple datasets with various metric learning methods to validate the effectiveness of our LSD.

\subsubsection{Ablation Studies on Different Loss Functions.} 
To validate the effectiveness of our method, we chose various popular loss functions of metric learning to conduct our experiments. Based on their definitions, they can be classified into 3 categories: pair-based, proxy-based and AP-based. We selected several representative methods in each category for our experiments:
\noindent \textbf{Pair-based:} Contrastive loss~\cite{hadsell2006dimensionality} and triplet loss~\cite{schroff2015facenet}.
\noindent \textbf{Proxy-based:} ProxyNCA loss~\cite{movshovitz2017no}. 
\noindent \textbf{AP-based:} Smooth-AP~\cite{brown2020smooth}. Moreover, to show realistic benefit, we also apply LSD to best performing objectives evaluated \cite{roth2020revisiting}, namely (i) Margin loss with Distance-based Sampling~\cite{wu2017sampling} and (ii) Multisimilarity loss~\cite{wang2019multi}.

Our LSD can be directly applied to the DML framework. We evaluate it with the above loss functions. Table~\ref{tab:results} shows the results. Note that the ‘+LSD’ notation means that the indicated model was trained with LSD and the related loss function. As shown in Table~\ref{tab:results}, in most cases, LSD significantly and consistently improves the original DML approaches on all benchmarks. 

\subsubsection{Ablation Studies on Different Sampling Strategies.} 
The data sampling strategy has been widely studied as an important part of metric learning. Most sampling strategies have focused on how to select useful samples from the training set. To analyze the impact of data sampling on LSD, we employ four popular data sampling strategies, random, semi-hard sampling~\cite{schroff2015facenet}, soft-hard mining~\cite{roth2019mic} and distance-based tuple mining~\cite{wu2017sampling}, with different loss functions in our experiments, respectively. The suffixes `(R)', `(H)', `(S)' and `(D)' denote the random, semi-hard, soft-hard and distance-based sampling strategy, respectively. As shown in Table~\ref{tab:results}, LSD can improve generalization through different sampling strategies. 

\subsubsection{Effect of Soft Embedding}
\label{subsubsec:soft embedding}
Although the previous experiments showed the effectiveness of LSD, since LSD is based on listwise loss, a question arises: does the improvement come not from \textit{soft embedding} but from listwise loss? In order to verify that the improvement of the performance comes from \textit{soft embedding} rather than from the use of multiple loss functions, we conducted the following experiments: we replaced the $s_{ij}^{(T)}$ of $R_{\textrm{LSD}}$ (Eq~\ref{eq:lsd2}) with hard label $y_i$ ($y_i = 0/1$ for negative/positive pairs) in the objective function (Eq~\ref{eq:lsd3}), i.e., we use hard labels instead of the similarities predicted from the teacher model as the objective targets. The image retrieval performance are shown in Table~\ref{tab:softmapping}. The results clearly indicate that the improvement comes from the \textit{soft embedding} rather than the combination of loss functions.

\tabcolsep=3pt
\begin{table}
\caption{Comparison of using \textit{soft embedding} and \textit{hard embedding} in LSD. }
\centering
\resizebox{\linewidth}{!}{
    \begin{tabular}{ccccccc}
    \toprule
    Benchmarks$\rightarrow$ & \multicolumn{2}{c}{CUB200-2011} & \multicolumn{2}{c}{CARS196}& \multicolumn{2}{c}{SOP}\\
    \midrule
    Approaches$\downarrow$ & \textbf{R@1} & \textbf{mAP} & \textbf{R@1} & \textbf{mAP} & \textbf{R@1} & \textbf{mAP}\\
    \midrule
    Triplet(H)           & 59.67& 31.13& 71.93& 29.58& 73.57& 41.17 \\
    \multirow{2}{*}{Triplet(H)+LSD w/ \textit{hard}}
                        & 54.10& 24.86&48.65& 19.61 & 62.83&30.61\\
                         &($\downarrow$5.57)&($\downarrow$6.27)& ($\downarrow$23.28)&($\downarrow$9.97) &  ($\downarrow$10.74)&  ($\downarrow$10.56)\\
    \multirow{2}{*}{Triplet(H)+LSD w/ \textit{soft}}
                        & \textbf{60.92}&  \textbf{32.67}& \textbf{75.05}&  \textbf{31.79}& \textbf{74.70}&  \textbf{42.11}\\
                        & \textbf{($\uparrow$ 1.25)}&  \textbf{($\uparrow$ 1.54)}& \textbf{($\uparrow$ 3.12)}&  \textbf{($\uparrow$ 2.21)}& \textbf{($\uparrow$ 1.13)}&  \textbf{($\uparrow$ 0.94)}\\
         
    \bottomrule
    \end{tabular}
}

\label{tab:softmapping}
\end{table}

\subsubsection{Ablation Studies on the weighting factor $\lambda$}
$\lambda$ is adopted to balance the metric learning loss and the knowledge distillation term in Eq~\ref{eq:lsd3} in our paper. To investigate the effect of the weighting factor $\lambda$, we provide the validation performances in terms of Recall Top-1 (R@1) on CUB200-2011, CARS196 and Stanford Online Products datasets with triplet loss. As demonstrated in Table~\ref{tab:lambda} the performance is consistent when changing $\lambda$ from 1.0 to 1000.0, with peaks around $\lambda=500$, $\lambda=75$ and $\lambda=100$ for CUB200-2011, CARS196 and Stanford Online Products.

\tabcolsep=15pt
\begin{table}
\caption{Ablation study on the value of the weighting factor $\lambda$ in Eq7 of our paper. We report the results on the three datasets with triplet loss.}
\centering
\resizebox{\linewidth}{!}{
    \begin{tabular}{cccc} 
    \hline  
    $\lambda$ & R@1 CUB200 & R@1 CARS196 & R@1 SOP \\
     \hline 
     										
     2      &   59.01&  73.29&   73.77\\
     5      &   59.37&  73.80&   74.04\\
     10      &   58.91&  73.13&   74.43\\
     20      &   59.82&  74.31&   74.55\\
     50      &   60.3&  74.54&   74.25\\
     75      &   -&  \textbf{75.05}&   -\\
     100      &   59.89&  74.27&   \textbf{74.65}\\
     200      &   60.25&  73.12&   74.46\\
     500      &   \textbf{60.92}&  72.03&   73.63\\
     1000      &   60.08&  69.59&   72.63\\
     \bottomrule
    \end{tabular} 
}

\label{tab:lambda}
\end{table}

\subsubsection{Effect of Temperature $\tau$}
To investigate the effect of temperature $\tau$ (in Eq~\ref{eq:lsd2}), we conduct the ablation experiments. As illustrated in Table~\ref{tab:tau}, we study the effect of temperature on Recall Top-1 (R@1) performances of LSD for CUB200-2011, CARS196 and Stanford Online Products datasets. We change the temperatures $\tau \in [0.05, 8] $. We observe that LSD is not sensitive to the temperature when $\tau \geq 1$ and hence set $\tau = 1$ by default.

\tabcolsep=15pt
\begin{table}
\caption{Ablation studies on the value of the temperature $\tau$ in Eq~\ref{eq:lsd}. We report the results on the three datasets with triplet loss.}
\centering
\resizebox{\linewidth}{!}{
    \begin{tabular}{cccc} 
    \hline  
    $\tau$ & R@1 CUB200 & R@1 CARS196 & R@1 SOP \\
     \hline 
     0.05   &   58.87&   72.72&   73.66\\
     0.1    &   59.21&   73.35&   74.13\\
     0.5    &   59.65&   74.26&   74.54\\
     1      &   \textbf{60.92}&   \textbf{75.05}&   74.65\\
     2      &   60.48&   74.92&   \textbf{74.74}\\
     4      &   60.71&   74.71&   74.63\\
     8      &   60.59&   74.35&   74.71\\
     \bottomrule
    \end{tabular} 
}

\label{tab:tau}
\end{table}

\subsubsection{Batchsize}
Batchsize determines the nature of the gradient updates to the network, \textit{e.g.} datasets with many classes benefit from large batchsizes due to better approximations of the training distribution. However, it is commonly not taken into account as a influential factor of variation~\cite{roth2020revisiting}. Our all experiments are based on the training protocol in~\cite{roth2020revisiting}, therefore we do not discuss more about batchsize.

\subsection{Comparison with Self-distillation Methods}
In this section, we compare LSD with the recent self-distillation methods, PS-KD~\cite{kim2021self}, CS-KD~\cite{yun2020regularizing} and Pointwise BAR~\cite{qin2021born}. PS-KD and CS-KD are regularizations proposed for classification tasks. Pointwise BAR is an alternative KD function for ranking tasks. In our experiments, we applied CS-KD and PS-KD based on cross entropy loss, and Pointwise BAR based on triplet loss. The comparison results are summarized in Table~\ref{tab:results}. We observe that training with LSD performs better than these state-of-the-art self-distillation methods. PS-KD and CS-KD are not applicable to metric learning, and, as concluded by \cite{qin2021born}, pointwise KD does not work well on ranking task. 

Here, we would like to explain why we did not compare with the SOTA method S2SD~\cite{roth2021simultaneous}. The motivation of S2SD is to use distillation to transfer information from high-dimensional features to low-dimensional features for the purpose of feature compression. This is different from our task. In addition, S2SD needs to modify the model structure and requires the model to output different high-dimensional features. The purpose of this paper is to demonstrate the effectiveness of self-distillation in the same dimension, so we only compare LSD with other self-distillation methods applied in a single dimensional space.

\begin{figure}[t]
\centering
\includegraphics[width=1.0\linewidth]{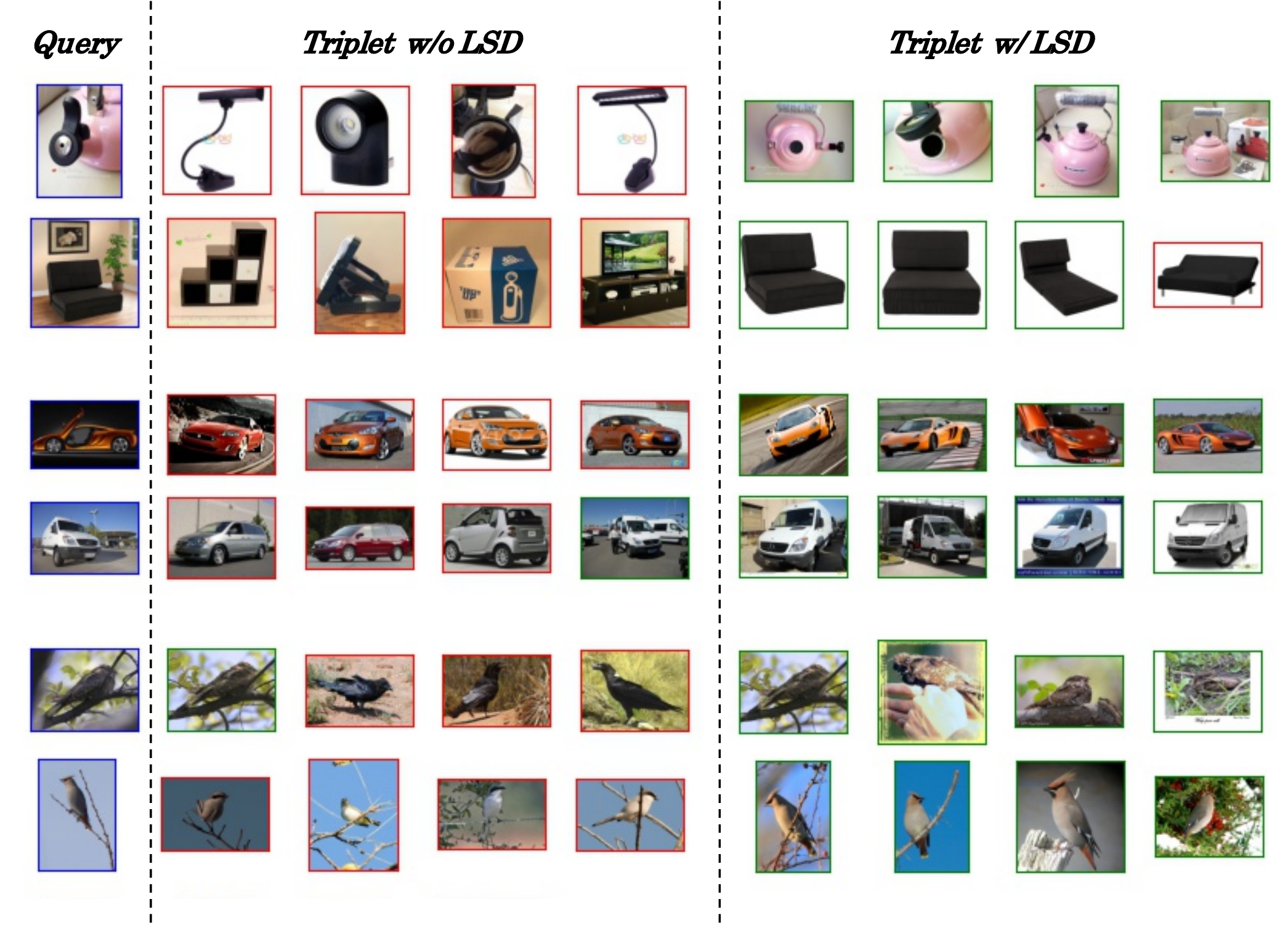}
\caption{The visualization of top 4 retrieved images w/o and w/ LSD. Green and red borders indicate correct and incorrect retrieved results, respectively.}
\label{fig:visiualization}
\end{figure}

\subsection{Domain Adaption}
For some image retrieval tasks, \textit{e.g.}, person re-id, domain adaption is an important task that corresponds to the capability of the model to transfer and preserve discriminative knowledge over multiple training sets. In other words, transfer learning is a good metric for assessing the generalizability of the model. In this section, we evaluate the ability of the proposed method to transfer ReID knowledge between old and new domains. We chose two of the most popular person re-id datasets, Market-1501 and Duke-MTMC, to conduct the experiments. We trained the model on one of the datasets using triplet loss, and then tested it on the other dataset. Table~\ref{tab:transfer} shows the results. We found that LSD significantly improved the performance of transfer learning.

\tabcolsep=5pt
\begin{table}
\caption{Results of domain adaption in the person re-id task.}
\centering
\resizebox{\linewidth}{!}{
    \begin{tabular}{ccccc}
    \toprule
    Benchmarks$\rightarrow$ & \multicolumn{2}{c}{Market to Duke} & \multicolumn{2}{c}{Duke to Market}\\
    \midrule
    Approaches$\downarrow$ & \textbf{R@1} & \textbf{mAP} & \textbf{R@1} & \textbf{mAP} \\
    \midrule
    Triplet(H)                 & 30.16& 16.55& 38.33& 16.21 \\
    \textbf{Triplet(H) + LSD}  & \textbf{34.39($\uparrow$4.23)}& \textbf{20.00($\uparrow$3.45)}& \textbf{44.61($\uparrow$6.28)}& \textbf{18.98($\uparrow$2.77)} \\ 
    \bottomrule
    \end{tabular}
}

\label{tab:transfer}
\end{table}

\subsection{Robustness to Noisy Data}
In this section, we demonstrate the robustness of LSD to noisy data. First, we generated noisy data on the three datasets (CUB200, CARS196 and d Stanford Online Products) with $\{0.1, 0.2, 0.3, 0.4, 0.5\}$ symmetric noisy ratios, following \cite{han2018co} and \cite{van2015learning}. Then we train models on the noisy data by using baseline (Triplet loss with semihard sampling strategy) and our proposed LSD. As reported in Table~\ref{tab:noisy}, the results indicate that LSD always achieves performance promotions on noisy training data, and this effect becomes more pronounced as the noisy ratio increases. This is in line with the conclusions of our analysis in Section~\ref{sec:theoretical}.




\tabcolsep=10pt
\begin{table}
\caption{Accuracy($\%$) on the three datasets validation with different noisy ratios on the training sets. }
\centering
\resizebox{\linewidth}{!}{
    \begin{tabular}{ccccc}
    \toprule
    \multirow{2}{*}{Noisy ratio} & Benchmarks$\rightarrow$ & CUB200 & CARS196& SOP\\
    &Approaches$\downarrow$ & \textbf{R@1} &  \textbf{R@1} &  \textbf{R@1} \\
    \midrule
    \multirow{2}{*}{0.1} & Triplet(H)           & 56.92& 67.79& 70.14 \\
    & Triplet(H)+LSD           & \textbf{58.95 ($\uparrow$ 2.03)}& \textbf{69.66 ($\uparrow$ 2.87)}& \textbf{74.41 ($\uparrow$ 1.27)}\\
    \midrule
    \multirow{2}{*}{0.2} & Triplet(H)           & 55.12& 63.88& 67.74 \\
    & Triplet(H)+LSD           & \textbf{58.30 ($\uparrow$ 3.18)}& \textbf{67.68 ($\uparrow$ 3.80)}& \textbf{69.85 ($\uparrow$ 2.21)}\\
    \midrule
    \multirow{2}{*}{0.3} & Triplet(H)           & 54.02& 57.74& 66.07 \\
    & Triplet(H)+LSD           & \textbf{56.93 ($\uparrow$ 2.91)}& \textbf{62.04 ($\uparrow$ 4.30)}& \textbf{67.50 ($\uparrow$ 2.43)}\\
    \midrule
    \multirow{2}{*}{0.4} & Triplet(H)           & 51.04& 49.62& 62.06 \\
    & Triplet(H)+LSD           & \textbf{55.44 ($\uparrow$ 4.40)}& \textbf{54.58 ($\uparrow$ 4.96)}& \textbf{64.91 ($\uparrow$ 2.85)}\\

    \bottomrule
    \end{tabular}
}

\label{tab:noisy}
\end{table}




\tabcolsep=10pt
\begin{table}[t]
\caption{Results of embedding space density ($\pi_{\text {ratio}}$).}
\centering
\resizebox{\linewidth}{!}{
    \begin{tabular}{l|c|c}
    \toprule
    Benchmarks$\rightarrow$ & {CARS196} & {SOP} \\
    \midrule
    Approaches$\downarrow$ & \multicolumn{2}{c}{$\pi_{\text {ratio}}$}\\
    \midrule
    Triplet (R)             
    &   0.2339&   0.2961\\
    Triplet (R) + LSD (Ours)
    &   \textbf{0.2876($\uparrow$0.0537)}&   \textbf{0.3754($\uparrow$0.0790)}\\
    Triplet (H)
    &   0.1856&   0.2805\\
    Triplet (H) + LSD (Ours)
    &   \textbf{0.2944($\uparrow$0.1088)}&   \textbf{0.3772($\uparrow$0.0967)}\\
    Triplet (D)
    &   0.2221&   0.3160\\
    Triplet (D) + LSD (Ours)
    &   \textbf{0.3321($\uparrow$0.1100)}&   \textbf{0.3451($\uparrow$0.0291)}\\
    \bottomrule
    \end{tabular}}

\label{tab:ratio}
\end{table}

\section{Embedding Space Metrics}
In this section, we investigate the effect of LSD on embedding space.

\subsection{Embedding Space Density}
Following \cite{roth2020revisiting}, we define the embedding space density as $\pi_{\text {ratio }}(\Phi)=\pi_{\text {intra }}(\Phi) / \pi_{\text {inter }}(\Phi)$, where average inter-class distances $\pi_{\textit{inter}}\left(\Phi\right)$ is,

\begin{equation}
\small
    \pi_{\text{inter}}\left(\Phi\right) = \frac{1}{Z_{\text {inter }}} \sum_{\substack{y_{l}, y_{k} \\ l \neq k}} d\left(\mu\left(\Phi_{y_{l}}\right), \mu\left(\Phi_{y_{k}}\right)\right)
\label{eq:inter}
\end{equation}
and average intra-class distances $\pi_{\text {intra }}(\Phi)$ is,

\begin{equation}
\small
    \pi_{\text {intra }}(\Phi)=\frac{1}{Z_{\text {intra }}} \sum_{y_{l} \in \mathcal{Y}} \sum_{\substack{\boldsymbol{v}_{i}, \boldsymbol{v}_{j} \in \Phi_{y_{l}}\\ i \neq j}} d\left(\boldsymbol{v}_{i}, \boldsymbol{v}_{j}\right)
\label{eq:intra}
\end{equation}
Here, $\Phi_{y_{l}}=\left\{\boldsymbol{v}_{i}:=\left \|f\left(\boldsymbol{x}_{i},\theta\right) \right \|_{2}\mid \boldsymbol{x}_{i} \in\mathcal{X}, y_{i}=y_{l}\right\}$ denotes the set of embedded samples of a class $y_{l}$, $\mu\left(\Phi_{y_{l}}\right)$ denotes their mean embedding and $Z_{\text {intra }}$ and $Z_{\text {inter }}$ are normalization constants. Table~\ref{tab:ratio} shows some results of embedding space density. We found that the embedding space density increase under the regularization of LSD, \textit{i.e.}, intra-class compactness in the embedding space decreases as expected under the regularization of LSD and the model performs significantly better than those without using LSD. These results support our hypothesis that learning a smoother embedding space helps improve the generalization of metric learning, and also consistent with the conclusion in \cite{roth2020revisiting} that an increased embedding space density ($\pi_{\text{ratio}}$) is linked to stronger generalisation. 

\subsection{Spectral Decay}
Following \cite{roth2020revisiting}, we analyze the spectral decay of the embedded training data $\boldsymbol{v}_{i}:=\{\left \|f\left(\boldsymbol{x}_{i},\theta\right) \right \|_{2}\mid \boldsymbol{x}_{i} \in\mathcal{X}\}$ via Singular Value Decomposition (SVD). Table~\ref{tab:decay} shows that, in most cases, the application of LSD lowers the spectral decay (thus providing a more feature diverse embedding space) across criteria, which is aligned with properties of improved generalization in DML as noted in \cite{roth2020revisiting}.

\tabcolsep=10pt
\begin{table}[t]
\caption{Results of spectural decay}
\centering
\resizebox{\linewidth}{!}{
    \begin{tabular}{l|c|c}
    \toprule
    Benchmarks$\rightarrow$ & {CARS196} & {SOP} \\
    \midrule
    Approaches$\downarrow$ & \multicolumn{2}{c}{Spectral Decay}\\
    \midrule
    Triplet (R)             
    &   \textbf{0.6442}&   0.7279\\
    Triplet (R) + LSD (Ours)
    &   0.6647($\uparrow$0.0205)&   \textbf{0.5875($\uparrow$0.1404)}\\
    Triplet (H)
    &   0.4602&   0.3251\\
    Triplet (H) + LSD (Ours)
    &   \textbf{0.4379($\downarrow$0.0223)}&   \textbf{0.2920($\downarrow$0
    0.0331)}\\
    Triplet (D)
    &   0.3951&   0.2084\\
    Triplet (D) + LSD (Ours)
    &   \textbf{0.3888($\downarrow$0.0062)}&    \textbf{0.1953($\downarrow$0.0131)}\\
    \bottomrule
    \end{tabular}}

\label{tab:decay}
\end{table}

\section{Visualization Results}
In this subsection, we visualize the retrieval results on the test set of the three datasets, respectively. Given a query image, we display its top 4 retrieved gallery images. The visualization results are shown in Fig.~\ref{fig:visiualization}. We found that our LSD facilitated the mining of fine-grained information. For example, in the last row, the model trained with LSD realized that the bird's crown feathers were a distinguishing feature of the query and retrieved the correct images. We provide more visualization results in the appendix.

\vspace{-3mm}
\section{Conclusion}
We propose a simple yet effective regularization, namely LSD, to improve the generalization performance of metric learning task. Instead of hard embedding, our LSD adopts soft embedding strategy. Specifically, LSD distills a model's own knowledge to assign more appropriate and informative targets for training. We also provide theoretical support, which shows that our LSD implicitly performs hard example mining and deeper information mining during training. From the experimental results conducted with various metric learning methods on multiple datasets, we observe that the proposed method is effective to improve the generalization capability of most metric learning methods.

\bibliographystyle{IEEEtran}
\bibliography{egbib}

\appendix
\subsection{Gradient Derivation}
\label{appendix_gradient}

In this section, we provide a full derivation of the loss gradient of LSD . For each anchor $\boldsymbol{x}_i$, LSD function can be described as:
\begin{equation}
\small
    R_{\textup{LSD}}(\mathcal{B}) = \frac{-1}{\left | \mathcal{B} \right |^2}
    \displaystyle\sum_{i \in \mathcal{B}^{\textrm{ID}}}
    \displaystyle\sum_{j\in\mathcal{B}^{\textrm{ID}}}\frac{t}{T}\sigma(s_{ij}^{(T)}/\tau)\textup{ln}(\sigma(s_{ij}^{(S)}/\tau)),
\label{appeq:lsd}
\end{equation}
For convenience, let's define:
\begin{equation}
\small
    \sigma(s_{ij}^{(T)}/\tau) = S_{ij},
\end{equation}

\begin{equation}
\small
    s_{ij}^{(t)} = \boldsymbol{z}_i \cdot \boldsymbol{z}_j,
\end{equation}
Then, LSD function can be described as:
\begin{equation}
\small
    \begin{split}
        R_{LSD} =  
          \frac{-t}{T \left | \mathcal{B} \right |^2}\sum_{\substack{j\in\mathcal{B}^{\textrm{ID}}}}S_{ij}\textup{ln}\frac{\textup{exp}(\boldsymbol{z}_i \cdot \boldsymbol{z}_j/\tau)}{\sum_{\substack{k\in\mathcal{B}^{\textrm{ID}}}}\textup{exp}(\boldsymbol{z}_i \cdot \boldsymbol{z}_k/\tau)},
    \end{split}
\label{appeq:lsd_convi}
\end{equation}
We now divide the gradient of Eq~\ref{appeq:lsd_convi} with respect to $\boldsymbol{v}_i$:
\begin{equation}
\small
    \frac{\partial R_{LSD}}{\partial \boldsymbol{v}_i} = \frac{\partial R_{LSD}}{\partial \boldsymbol{z}_i} \frac{\partial \boldsymbol{z}_i}{\partial \boldsymbol{v}_i},
\end{equation}
For the $\frac{\partial \boldsymbol{z}_i}{\partial \boldsymbol{v}_i}$, since $\boldsymbol{z}_i = \frac{\boldsymbol{v}_i}{\left \| \boldsymbol{v}_i \right \|}$, we can deduce that:
\begin{equation}
\small
    \begin{split}
        \frac{\partial \boldsymbol{z}_i}{\partial \boldsymbol{v}_i}
        &= \frac{\partial}{\partial \boldsymbol{v}_i} \left ( \frac{\boldsymbol{v}_i}{\left \| \boldsymbol{v}_i \right \|} \right )
        \\
        &=\frac{1}{\left \| \boldsymbol{v}_i \right \|} \textbf{I} - \frac{ \boldsymbol{v}_i \cdot \boldsymbol{v}_i  }{\left \| \boldsymbol{v}_i \right \|}
        \\
        &=\frac{1}{\left \| \boldsymbol{v}_i \right \|} \left (\textbf{I} - \frac{ \boldsymbol{v}_i \cdot \boldsymbol{v}_i  }{\left \| \boldsymbol{v}_i \right \|^2}\right )
        \\
        &=\frac{1}{\left \| \boldsymbol{v}_i \right \|} \left (\textbf{I} - \boldsymbol{z}_i \cdot \boldsymbol{z}_i\right ),
    \end{split}
    \label{appeq:lsd_grad_z2v}
\end{equation}

For the gradient with respect to $\boldsymbol{z}_i$:
\begin{equation}
    \scriptsize
    \begin{split}
        &\frac{\partial R_{LSD}}{\partial \boldsymbol{z}_i}
        \\
        &=\frac{-t}{T \left | \mathcal{B}\right |^2}\sum_{\substack{j \in \mathcal{B}^{\textrm{ID}}}}\frac{\partial}{\partial \boldsymbol{z}_i}S_{ij} \left \{ (\frac{\boldsymbol{z}_i \cdot \boldsymbol{z}_j}{\tau})-\textup{ln} \sum_{\substack{k \in \mathcal{B}^{\textrm{ID}}}}\textup{exp}\left( \frac{\boldsymbol{z}_i \cdot \boldsymbol{z}_k}{\tau} \right) \right \}
        \\
        &=\frac{-t}{\tau T \left | \mathcal{B} \right |^2}\sum_{\substack{j \in \mathcal{B}^{\textrm{ID}}}}\left \{ S_{ij} \left [ \boldsymbol{w}_{j} \boldsymbol{z}_j -\frac{\sum_{\substack{k \in \mathcal{B}^{\textrm{ID}}}} \boldsymbol{w}_{k}\boldsymbol{z}_k \textup{exp} \left( \frac{\boldsymbol{z}_i \cdot \boldsymbol{z}_k}{\tau} \right)}{\sum_{\substack{k \in \mathcal{B}^{\textrm{ID}}}}\textup{exp} \left( \frac{\boldsymbol{z}_i \cdot \boldsymbol{z}_k}{\tau} \right)}  \right ] \right \}
        \\
        &=\frac{-t}{\tau T \left | \mathcal{B} \right |^2}\left \{\sum_{\substack{j \in \mathcal{B}^{\textrm{ID}}}} S_{ij} \boldsymbol{w}_{j}\boldsymbol{z}_j - \sum_{\substack{j \in \mathcal{B}^{\textrm{ID}}}}S_{ij} \sum_{\substack{k \in \mathcal{B}^{\textrm{ID}}}} \boldsymbol{w}_{k}\boldsymbol{z}_k P_{ik}
        \right \}
        \\
        &=\frac{-t}{\tau T \left | \mathcal{B} \right |^2}\left \{\sum_{\substack{j \in \mathcal{B}^{\textrm{ID}}}} S_{ij} \boldsymbol{w}_{j}\boldsymbol{z}_j - \sum_{\substack{k \in \mathcal{B}^{\textrm{ID}}}} P_{ik} \boldsymbol{w}_{k}\boldsymbol{z}_k 
        \right \}
        \\
        &=\frac{-t}{\tau T \left | \mathcal{B} \right |^2}\left \{\sum_{\substack{j \in \mathcal{B}^{\textrm{ID}}}} \boldsymbol{w}_{j}\boldsymbol{z}_j \left (  S_{ij}-P_{ij}
        \right ) 
        \right \}
        \\
        &=\frac{-t}{\tau T \left | \mathcal{B} \right |^2}\left \{\sum_{p \in \mathcal{P}^{*}(i)} \!\! \boldsymbol{w}_{p}\boldsymbol{z}_p \left (  S_{ip}\!- \!P_{ip}\right ) \!+
        \!\!\!\!\!\sum_{n \in \mathcal{N}(i)} \!\!\boldsymbol{w}_{n}\boldsymbol{z}_n \left (  S_{in}\!- \!P_{in}\right ) 
        \right \}
    \end{split}
\label{appeq:lsd_grad_l2z}
\end{equation}
where:
\begin{equation}
\small
    \begin{split}
        P_{ik} = \frac{ \textup{exp} \left( \frac{\boldsymbol{z}_i \cdot \boldsymbol{z}_k}{\tau} \right)}{\sum_{k \in \mathcal{B}^{\textrm{ID}}}\textup{exp} \left( \frac{\boldsymbol{z}_i \cdot \boldsymbol{z}_k}{\tau} \right)} 
    \end{split}
\end{equation}
and
\begin{equation}
\small
    \boldsymbol{w}_{j}=\begin{cases}
        1 & \text{ if } j \neq i \\ 
        2 & \text{ if } j = i,
    \end{cases}
\end{equation}
Here, $\mathcal{P}^{*}(i)$ and $\mathcal{N}(i)$ are the positive set (include the anchor $\boldsymbol{x}_{i}$ itself) and negative set of $\boldsymbol{x}_{i}$ in the mini-batch $\mathcal{B}$, respectively.
Combining Eq~\ref{appeq:lsd_grad_z2v} and Eq~\ref{appeq:lsd_grad_l2z} thus gives:
\begin{equation}
\small
    \begin{split}
        &\frac{\partial R_{LSD}}{\partial \boldsymbol{v}_i} 
        \\
        &= \frac{-t}{\tau T\left | \mathcal{B} \right |^2 \left \| \boldsymbol{v}_i \right \|}\left (\textbf{I} - \boldsymbol{z}_i \cdot \boldsymbol{z}_i\right )\left \{\sum_{p \in \mathcal{P}^{*}(i)} \boldsymbol{w}_{p} \boldsymbol{z}_p \left (  S_{ip}-P_{ip}\right ) \right. \\
        &\left. +
        \sum_{n \in \mathcal{N}(i)} \boldsymbol{w}_{n}\boldsymbol{z}_n \left (  S_{in}-P_{in}\right ) 
        \right \}
        \\
        &= \frac{t}{\tau T \left | \mathcal{B} \right |^2 \left \| \boldsymbol{v}_i \right \|} \left \{\sum_{p \in \mathcal{P}(i)} \left ( \boldsymbol{z}_p -\left ( \boldsymbol{z}_i \cdot \boldsymbol{z}_p \right ) \boldsymbol{z}_i\right ) \left ( P_{ip} - S_{ip}\right ) \right. \\
        &\left. +
        \sum_{n \in \mathcal{N}(i)} \left ( \boldsymbol{z}_n -\left ( \boldsymbol{z}_i \cdot \boldsymbol{z}_n \right ) \boldsymbol{z}_i\right )\left ( P_{in} - S_{in}\right ) 
        \right \} \\
        &= \left. \frac{\partial R_{LSD}}{\partial \boldsymbol{v}_i} \right|_{\mathcal{P}(i)} 
        + \left. \frac{\partial R_{LSD}}{\partial \boldsymbol{v}_i} \right|_{\mathcal{N}(i)} 
    \end{split}
\end{equation}
where:
\begin{equation}
    \small
    \begin{split}
    & \left. \frac{\partial R_{LSD}}{\partial \boldsymbol{v}_i} \right|_{\mathcal{P}(i)} \\
    & = \frac{t}{\tau T \left | \mathcal{B} \right |^2 \left \| \boldsymbol{v}_i \right \|}\sum_{p \in \mathcal{P}(i)} \left ( \boldsymbol{z}_p -\left ( \boldsymbol{z}_i \cdot \boldsymbol{z}_p \right ) \boldsymbol{z}_i\right ) \left (   P_{ip} - S_{ip}\right )
    \end{split}
\end{equation}
\begin{equation}
    \small
    \begin{split}
    & \left. \frac{\partial R_{LSD}}{\partial \boldsymbol{v}_i} \right|_{\mathcal{N}(i)} \\
    & = \frac{1}{\tau \left | \mathcal{B} \right |^2 \left \| \boldsymbol{v}_i \right \|}\sum_{n \in \mathcal{N}(i)} \left ( \boldsymbol{z}_n -\left ( \boldsymbol{z}_i \cdot \boldsymbol{z}_n \right ) \boldsymbol{z}_i\right ) \left (   P_{in}-S_{in}\right )
    \end{split}
\end{equation}

Here, $\mathcal{P}^(i)$ is the positive set not include the anchor $\boldsymbol{x}_{i}$.

\begin{figure}
\centering
\includegraphics[width=.95\linewidth]{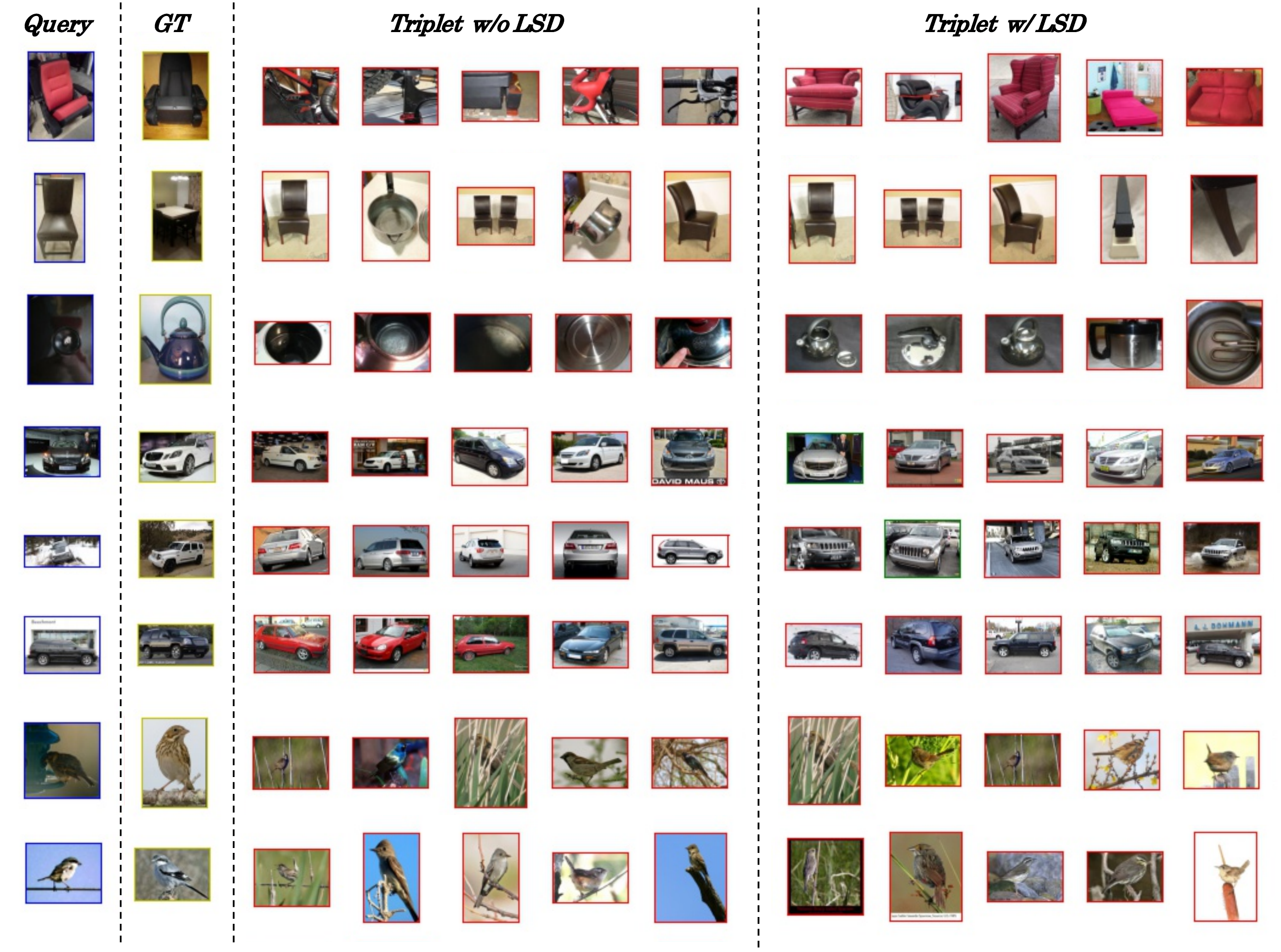}
\vspace{-2mm}
\caption{Visualization failure cases. Given a query image (blue), we present a ground truth sample (yellow) and the visualization of top 5 retrieved images w/o and w/ LSD. Green and red borders indicate correct and incorrect retrieved results, respectively.}
\label{fig:bad}
\end{figure}

In addition, we also present some bad cases in the Figure~\ref{fig:bad}, where our retrieved results are visually closer to the query than that of baseline model.

\begin{figure}
\centering
\includegraphics[width=.95\linewidth]{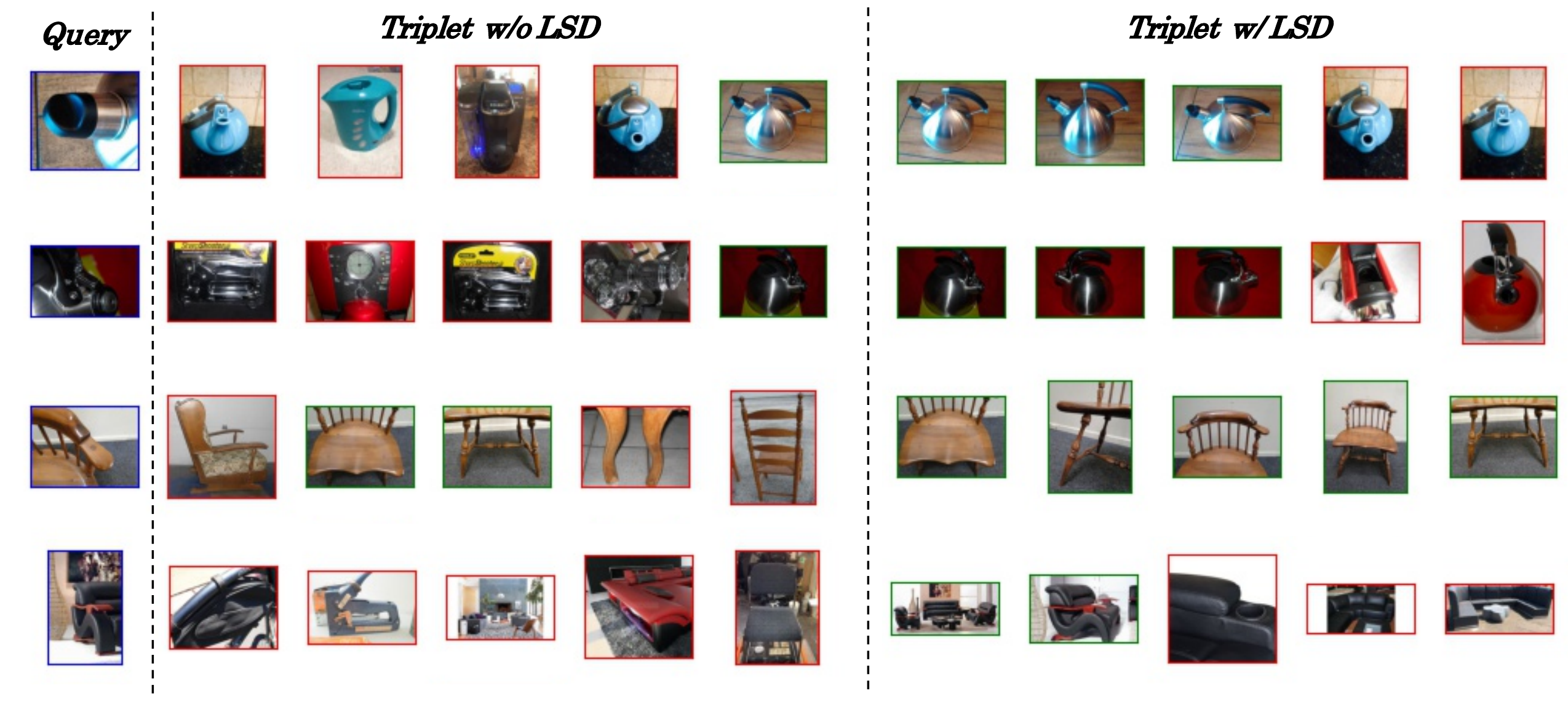}
\caption{Given a local query image (blue), we present the visualization of top 5 retrieved images w/o and w/ LSD. Green and red borders indicate correct and incorrect retrieved results, respectively.}
\label{fig:local2global}
\end{figure}

\begin{figure}
\centering
\includegraphics[width=.95\linewidth]{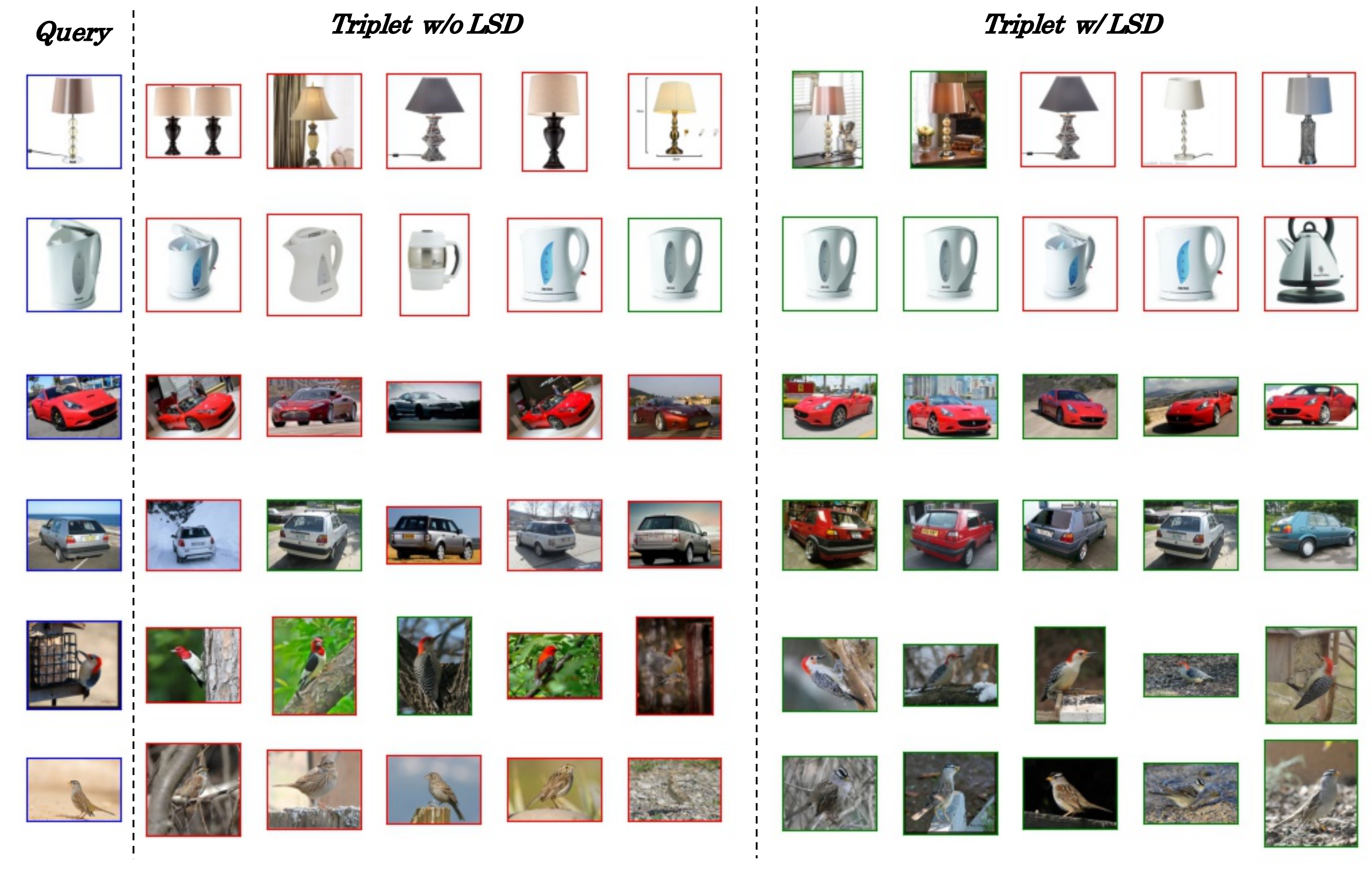}
\vspace{-2mm}
\caption{Given a query image (blue), we present the visualization of top 5 retrieved images w/o and w/ LSD. There are some negative samples in gallery that are very similar to the query image. Green and red borders indicate correct and incorrect retrieved results, respectively.}
\label{fig:detial}
\end{figure}

\begin{figure}
\centering
\includegraphics[width=.95\linewidth]{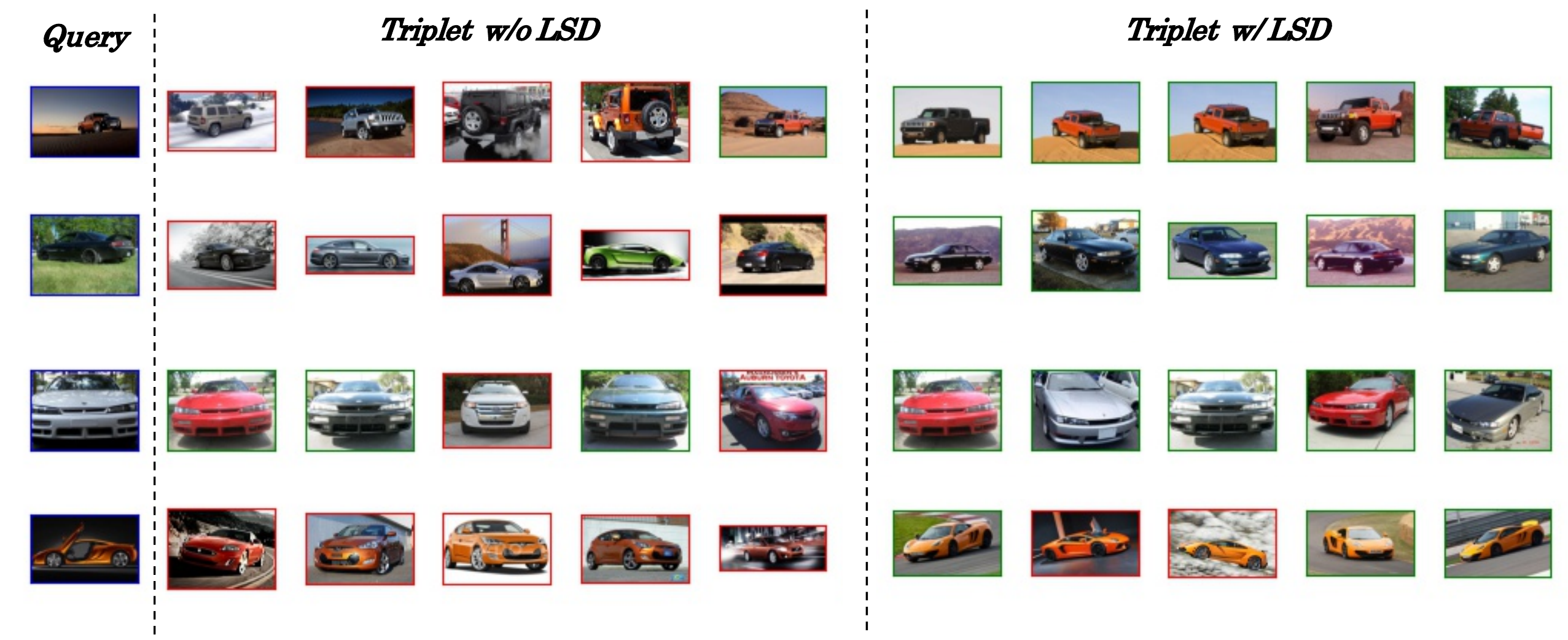}
\vspace{-2mm}
\caption{Visualization results of cross-view image retrieval. Given a query image (blue), we present the visualization of top 5 retrieved images w/o and w/ LSD. Green and red borders indicate correct and incorrect retrieved results, respectively.}
\label{fig:cross-view}
\end{figure}

\subsection{Visualization Results and Failure Case}
In this section, we visualize a number of samples to investigate the performance of our LSD. Specifically, we validate the capability of LSD in three aspects by visualizing the retrieval results: 
\begin{itemize}
    \item \textbf{Local-to-global}: The query image contains only local content of the target. Figure~\ref{fig:local2global} shows that LSD improve the model's capability to handle``local-to-global". For example, at the first row, our model is aware of the query image is the spout of a stainless steel kettle. 
    \item \textbf{Detail differences}: The negative samples are similar to the query image, and only the details differ. Figure~\ref{fig:detial} demonstrates the ability of our model to discriminate detail differences. For example, at the third row, our model observes the details of the air intake at the front of the car.
    \item \textbf{Cross-view}: The positive samples in the gallery have different views from the query image. Figure~\ref{fig:cross-view} shows that our model can work well on the cross-view task. 
\end{itemize}

\end{document}